\documentclass{article}

\usepackage[final]{neurips_2024}
\usepackage{fancyhdr} 
\usepackage{datetime2}

\fancypagestyle{firstpage}{
    \fancyhf{} 
    \fancyhead[L]{Babel} 
\fancyhead[C]{}         
\fancyhead[R]{Feb. 2025} 
}

\fancyhfoffset[L]{0pt} 
\fancyhfoffset[R]{0pt} 

\usepackage{array}
\usepackage{colortbl}
\usepackage{xcolor}

\usepackage{xcolor}
\definecolor{customblue}{RGB}{0, 0, 123}
\definecolor{lightblue}{rgb}{0.9, 0.95, 1.0}
\definecolor{almond}{rgb}{0.94, 0.87, 0.8}
\definecolor{amber}{rgb}{1.0, 0.75, 0.0}
\definecolor{alizarin}{rgb}{0.82, 0.1, 0.26}
\definecolor{antiquewhite}{rgb}{0.98, 0.92, 0.84}
\definecolor{codegray}{RGB}{220, 220, 220}
\newcolumntype{g}{>{\columncolor{codegray!25}}r}

\usepackage{hyperref} 
\hypersetup{
    colorlinks=true,       
    linkcolor=customblue,        
    urlcolor=customblue,        
    citecolor=customblue      
}

\usepackage{hyperref} 
\hypersetup{
    colorlinks=true,       
    linkcolor=customblue,        
    urlcolor=customblue,        
    citecolor=customblue      
}
\usepackage{amsmath}
\usepackage{adjustbox}
\usepackage[utf8]{inputenc} 
\usepackage[T1]{fontenc}    
\usepackage{hyperref}       
\usepackage{url}            
\usepackage{booktabs}       
\usepackage{amsfonts}       
\usepackage{nicefrac}       
\usepackage{microtype}      
\usepackage{xcolor}         
\usepackage{times}
\usepackage{latexsym}
\usepackage{amsmath}
\usepackage{amsfonts}
\usepackage{graphicx}
\usepackage{pifont}
\usepackage{graphicx}
\usepackage[T1]{fontenc}

\usepackage[utf8]{inputenc}
\usepackage{multirow}
\usepackage{microtype}
\usepackage{booktabs}
\usepackage{graphicx}
\usepackage{subcaption}
\usepackage{wrapfig}
\usepackage{calligra}
\usepackage{mathpazo}
\usepackage{yfonts}
\usepackage{tcolorbox}
\usepackage{lipsum}
\usepackage{tikz}
\usepackage{lipsum}
\usetikzlibrary{shadows.blur}
\usepackage{enumitem}

\title{
    \begin{center}
        \begin{minipage}{0.1\textwidth}
            \centering
            \includegraphics[height=40pt]{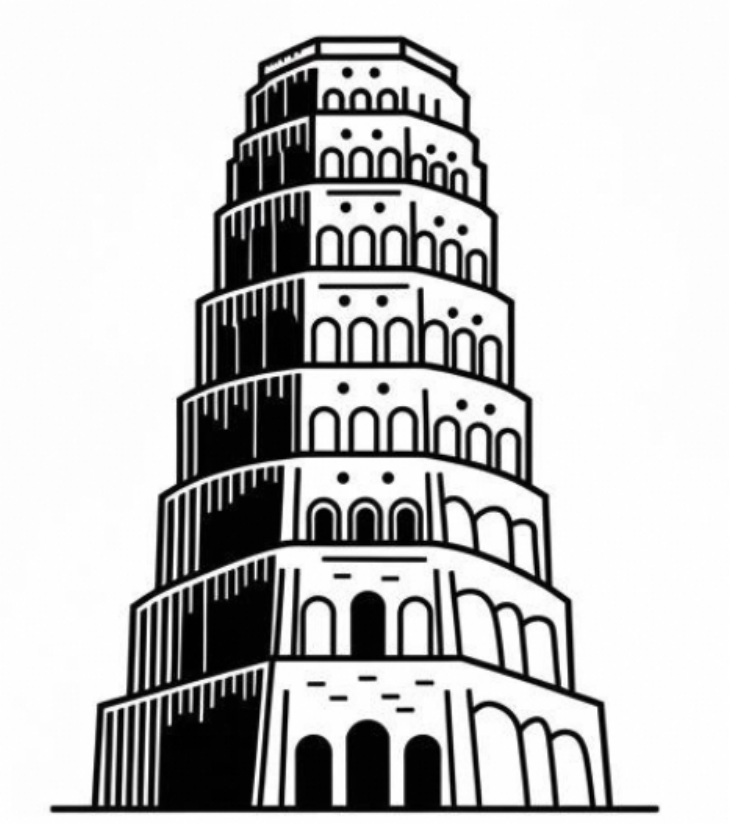}
        \end{minipage}%
        \begin{minipage}{0.9\textwidth}
            \centering
            Babel: Open Multilingual Large Language Models Serving Over
            90\% of Global Speakers
        \end{minipage}
    \end{center}
}

\author{
\hspace{-0.4cm} Yiran Zhao \; Chaoqun Liu \; Yue Deng \; Jiahao Ying \; Mahani Aljunied \;   \textbf{Zhaodonghui Li} \\
 \textbf{Lidong Bing} \; \textbf{Hou Pong Chan} \; \textbf{Yu Rong} \; \textbf{Deli Zhao} \; \textbf{Wenxuan Zhang}\footnotemark[2] \\
   \\
  \textbf{DAMO Academy, Alibaba Group} \\
\\
\hspace{-0.2cm}\texttt{{\{zhaoyiran.zyr,\;royrong.ry\}@alibaba-inc.com, wxzhang@sutd.edu.sg}}
  \\
  \\
Project page: \href{https://babel-llm.github.io/babel-llm/}{https://babel-llm.github.io/babel-llm/}
}
\begin{document}

\maketitle

\renewcommand{\thefootnote}{\fnsymbol{footnote}}
\footnotetext[2]{Wenxuan Zhang is the corresponding author.}
\renewcommand{\thefootnote}{\arabic{footnote}}

\vspace{-0.4cm}
\begin{center}
\begin{tikzpicture}
\node[
    draw=black, 
    fill=gray!5, 
    text width=0.8\linewidth, 
    align=center, 
    inner sep=10pt, 
    rounded corners=15pt, 
    drop shadow={fill=black!50, opacity=0.3} 
] 
{
\textit{\fontfamily{ppl}\selectfont 
People built the \textbf{Tower of Babel} to reach heaven and achieve unity, \\
but God confused their language and scattered them across the earth.} \\[1em]
-- \textit{Story from Genesis, Old Testament}
};
\end{tikzpicture}
\end{center}

\begin{abstract}

Large language models (LLMs) have revolutionized natural language processing (NLP), yet open-source multilingual LLMs remain scarce, with existing models often limited in language coverage. Such models typically prioritize well-resourced languages, while widely spoken but under-resourced languages are often overlooked. To address this disparity, we introduce \texttt{Babel}, an open multilingual LLM that covers the top 25 languages by number of speakers, supports over 90\% of the global population, and includes many languages neglected by other open multilingual LLMs. Unlike traditional continue pretraining approaches, Babel expands its parameter count through a layer extension technique that elevates Babel's performance ceiling. We introduce two variants: \texttt{Babel-9B}, designed for efficient inference and fine-tuning, and \texttt{Babel-83B}, which sets a new standard for open multilingual LLMs. Extensive evaluations on multilingual tasks demonstrate its superior performance compared to open LLMs of comparable size. In addition, using open-source supervised fine-tuning datasets, Babel achieves remarkable performance, with Babel-9B-Chat leading among 10B-sized LLMs and Babel-83B-Chat setting a new standard for multilingual tasks, reaching the same level of commercial models.

\end{abstract}

\section{Introduction}

Large language models (LLMs)~\citep{achiam2023gpt, reid2024gemini, dubey2024llama, team2024gemma} have revolutionized the field of natural language processing (NLP), emerging as powerful tools that drive innovation and improve various aspects of human life \citep{ li2023huatuo26m, roziere2023code, li2023large, yang2024qwen2, alves2024tower}. However, multilingual LLMs remain relatively rare, particularly in the open-source domain~\citep{hurst2024gpt, claude35sonnet}, where the supply of such models falls short of the growing demand for broader language support.
Even among the existing open-source multilingual LLMs, the range of languages they support is often constrained. These models, such as Bloom~\citep{workshop2023bloom176bparameteropenaccessmultilingual}, GLM-4~\citep{glm2024chatglm} and Qwen2.5~\citep{qwen2025qwen25technicalreport}, tend to prioritize languages with extensive training resources—typically those spoken in developed countries, such as French, Arabic, and German, where large volume of high-quality datasets are readily available. In contrast, languages spoken in less developed regions, such as Hindi, Bengali, and Urdu—despite having millions of speakers, often outnumbering those of French or German~\citep{eberhard2024ethnologue}—receive considerably less attention.


To bridge this gap and make LLMs more accessible to a wider global audience, we introduce \texttt{Babel} - a new open-source multilingual large language model, aiming to serve over $90$\% of speakers worldwide. Specifically, we focus on the top 25 languages by number of speakers, including English, Chinese, Hindi, Spanish, Arabic, French, Bengali, Portuguese, Russian, Urdu, Indonesian, German, Japanese, Swahili, Filipino, Tamil, Vietnamese, Turkish, Italian, Javanese, Korean, Hausa, Persian, Thai, and Burmese. Notably, more than half of these languages, despite being widely spoken, have been largely neglected by existing open-source multilingual LLMs. Furthermore, given the limited availability of high-quality training data for many of these languages, we place significant emphasis on optimizing the data-cleaning pipeline to ensure the highest possible data quality. To this end, we collect data from diverse sources and employ an LLM-based quality classifier to curate clean, high-quality content for training.

Unlike conventional continue pretraining approaches~\citep{seallms, zhao2024llama}, we improve Babel's performance ceiling by increasing its parameter space through model expansion. Specifically, we employ layer extension, a structured approach that adds new layers identical in architecture to the original ones. Balancing accessibility and state-of-the-art performance, we present two model variants: \texttt{Babel-9B} and \texttt{Babel-83B}. \texttt{Babel-9B} is designed for efficient multilingual LLM inference and fine-tuning, making it ideal for research and local deployment, while \texttt{Babel-83B} establishes a new benchmark as the leading open multilingual LLM. A comprehensive evaluation on multilingual datasets highlights Babel's superior performance compared to open LLMs of a similar size. Furthermore, we leverage open-source supervised fine-tuning (SFT) datasets including WildChat~\citep{zhaowildchat} and Everything Instruct Multilingual~\citep{everything_instruct_multilingual} to construct an SFT training pool of 1 million conversations without creating additional training data. This pool is then used to train Babel-9B-Base and Babel-83B-Base. We show that even with publicly available fine-tuning data, Babel chat models demonstrate strong task-solving capabilities. Notably, Babel-9B-Chat achieves state-of-the-art performance among 10B-sized LLMs, while Babel-83B-Chat sets a new benchmark for open LLMs and even performs comparably to state-of-the-art commercial models such as GPT-4o on certain tasks. These results highlight the strong foundational performance of Babel base models.



\section{Supported Languages}

Current LLMs increasingly support non-English languages such as GLM-4~\citep{glm2024chatglm} and Qwen2.5~\citep{qwen2025qwen25technicalreport}. However, these models primarily focus on languages with extensive training corpora, which are often spoken in developed countries, such as French and German, where numerous research institutions curate and process high-quality data. In contrast, languages spoken in less developed countries, such as Hindi (345 million L1 speakers), Bengali (237 million L1 speakers), and Urdu (70 million L1 speakers), receive comparatively less attention. For context, Spanish is spoken by 486 million L1 speakers, French by 74 million, and German by 76 million. 
To make LLMs more accessible to a broader audience, we selected languages based on the number of speakers. Specifically, we included a total of 25 languages, with detailed statistics provided in Table \ref{tab:languages}~\citep{eberhard2024ethnologue}. Altogether, \texttt{Babel} serves around 7 billion speakers globally, covering more than 90\% of the world’s population.


\begin{table*}[t]
\centering
\renewcommand{\arraystretch}{1.2}
\footnotesize
  \scalebox{1}{
\begin{tabular}{@{}lcllc@{}}
\toprule
\textbf{Language}         & \textbf{Speakers} & \textbf{Language Family} & \textbf{Macroarea}   &  \textbf{CC ratio}     \\ 
\midrule
English                   & 1.5B              & Germanic                 & Worldwide        &    43.4           \\
Chinese (Mandarin)        & 1.4B              & Sinitic                  & Asia                & 5.1           \\
\rowcolor{lightblue}Hindi                     & 700M              & Indo-Aryan               & Asia & 0.2                          \\
Spanish                   & 595M              & Romance                  & Americas, Europe    & 4.6           \\
Standard Arabic           & 400M              & Semitic                  & Asia, Africa   & 0.68                \\
French                    & 300M              & Romance                  & Europe, Africa, Americas & 4.4      \\
\rowcolor{lightblue}Bengali                   & 300M              & Indo-Aryan               & Asia  & 0.1                         \\
Portuguese                & 270M              & Romance                  & Americas, Europe, Africa   & 2.3   \\
Russian                   & 260M              & Slavic                   & Europe, Asia     & 6.2       \\
\rowcolor{lightblue}Urdu                      & 230M              & Indo-Aryan               & Asia  & 0.02                         \\
\rowcolor{lightblue}Indonesian                & 200M              & Malayo-Polynesian        & Asia  & 1.1                         \\
Standard German           & 135M              & Germanic                 & Europe        & 5.4                 \\
Japanese                  & 130M              & Japonic                  & Asia        & 5.3                   \\
\rowcolor{lightblue}Swahili                   & 100M              & Bantu                    & Africa  & 0.008                       \\
\rowcolor{lightblue}Filipino (Tagalog)        & 100M              & Malayo-Polynesian        & Asia   & 0.008                        \\
\rowcolor{lightblue}Tamil                     & 90M               & Dravidian                & Asia    & 0.04                       \\
Vietnamese                & 86M               & Vietic                   & Asia   & 1.0                        \\
Turkish                   & 85M               & Turkic                   & Asia, Europe        & 1.3           \\
Italian                   & 85M               & Romance                  & Europe         & 2.4               \\
\rowcolor{lightblue}Javanese                  & 83M               & Malayo-Polynesian        & Asia  & 0.002                         \\
Korean                    & 80M               & Koreanic                 & Asia         &    0.76              \\
\rowcolor{lightblue}Hausa                     & 80M               & Chadic                   & Africa & 0.003                        \\
\rowcolor{lightblue}Iranian Persian           & 80M               & Indo-Iranian             & Asia    & 0.74                       \\
\rowcolor{lightblue}Thai                      & 80M               & Kra-Dai                  & Asia   & 0.42                        \\
\rowcolor{lightblue}Burmese                   & 50M               & Tibeto-Burman            & Asia   & 0.01                        \\ 
\bottomrule
\end{tabular}}
\caption{Languages supported by Babel sorted by the number of speakers (B = Billion, M = Million). CC ratio indicates the number of open training corpora. Highlighted languages are those underexplored by previous multilingual LLMs. 
}
\label{tab:languages}
\end{table*}

\section{Data Preparation}

\subsection{Data Collection}

Building on the foundation of prior work \citep{team2024gemma, sailor, zhang2024seallms}, we have diversified the range of data sources.
In particular, we have incorporated essential knowledge from resources like Wikipedia \citep{wikidump} and textbooks \citep{benallal2024cosmopedia}, journalistic content from CC-News \citep{ccnews}, web-based corpora such as CulturaX \citep{nguyen2023culturax}, and the MADLAD-400 dataset \citep{kudugunta2023madlad400}.

\subsection{LLMs-based Data Cleaning and Processing} 

Due to the limited availability of high-quality training data for many of these languages, we place significant emphasis on optimizing the data-cleaning pipeline to ensure the highest possible data quality. The detailed procedures are outlined as follows.

\begin{enumerate}[label={(\arabic*)}]
    \item \textbf{Normalization.} 
    
    We apply predefined rules to filter out low-quality data, such as documents with fewer than 100 characters or those containing more than $30\%$ digits.

    \item \textbf{LLMs-based quality classifier.}

    We train the classifier based on the Qwen-2.5-0.5B-Instruct model~\citep{qwen2025qwen25technicalreport}, leveraging a method that combines strong model-based labeling with expert linguistic refinement to construct the training dataset. Specifically, we adopt the ``LLM-as-a-judge'' approach, where GPT-4o is prompted to evaluate and score potential training data across various dimensions. These initial scores are then carefully reviewed by linguistic experts to ensure that only high-quality data is selected for training the evaluator. 

    \item \textbf{Deduplicate.}

    We identify and remove duplicate documents by hashing, pairing duplicates, constructing graphs, and recording duplicates for removal.

\end{enumerate}





\section{Model Description}





\subsection{Model Extension}

\begin{wrapfigure}{r}{0.45\textwidth}
\vspace{-0.5cm}
    \centering
    \includegraphics[width=0.45\textwidth]{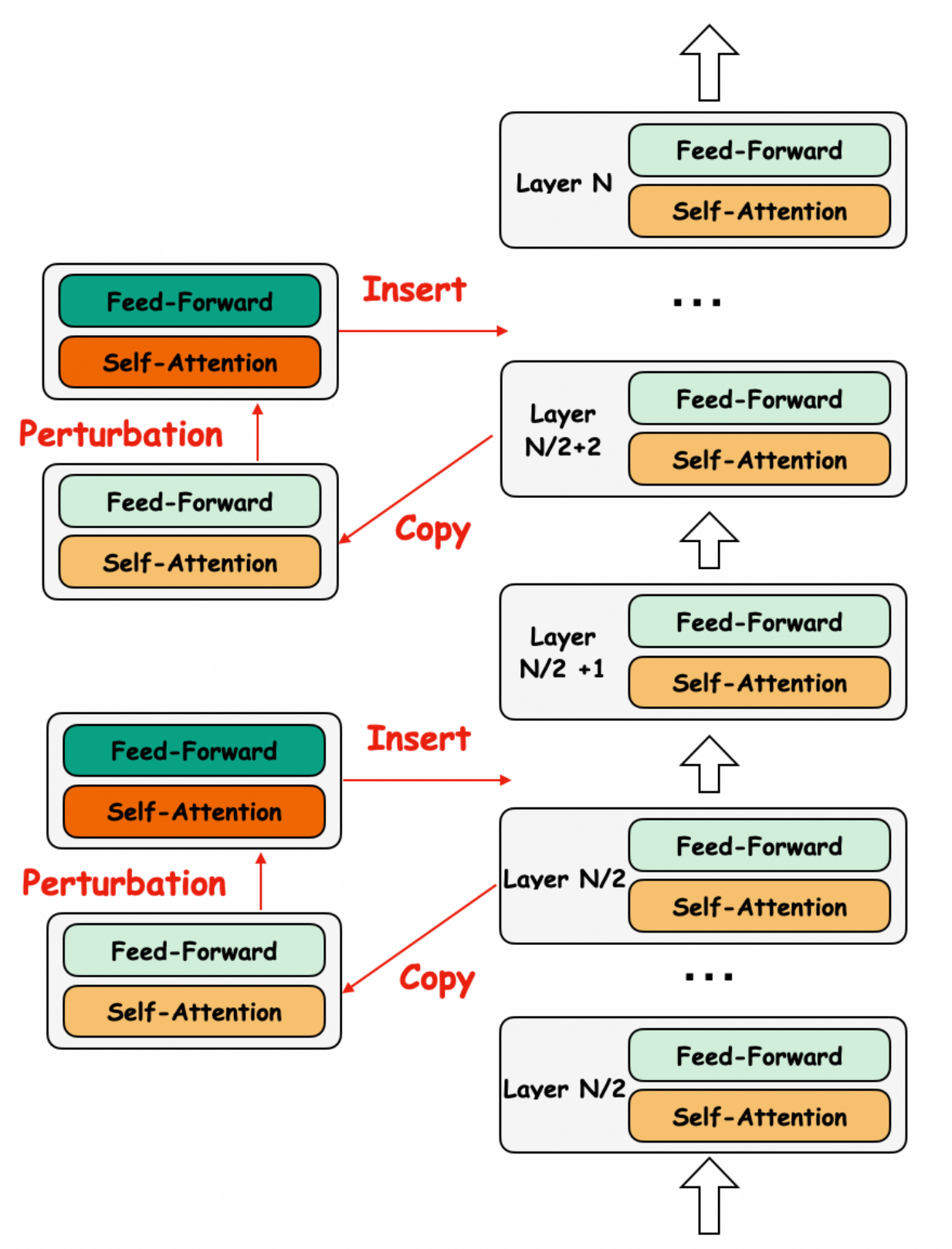}
   \caption{Layer extension for Babel. }
    \label{fig:layer}
\end{wrapfigure}

To improve the model's performance upper bound, we increase its parameter count through model expansion. Specifically, as illustrated in Figure \ref{fig:layer}, we use layer extension, a structured method that directly adds new layers with the same structure as the original ones. This approach does not affect critical components of the model, such as attention heads, hidden embeddings, or the embedding layer, etc. Furthermore, inspired by the observation that the middle and back layers are less sensitive to editing~\citep{kim2023solar, men2024shortgpt, zhang2024finercut}, we choose to extend the layers in the second half of the model. 

We explore various layer extension settings, including different positions for adding layers and methods for initialization of newly added parameters. Specifically, we experiment with inserting layers between the original layers or appending them after the original model. For parameter initialization, we consider duplicating the original parameters, initializing with all zero, or adding noise to the original parameters. We select the optimal layer extension methods based on the models' initial performance, prioritizing those that minimally affect performance while also considering their impact on further training.

\subsection{Analysis}

We conduct an ablation analysis of initialization methods using Qwen2.5-72B-Base~\citep{qwen2025qwen25technicalreport} as the backbone model and evaluate performance on the MMMLU~\citep{openai2024mmmlu}. Specifically, we examine two key aspects: (1) the layer insertion position, which can be either among existing layers or directly appended to the final layer of the original model, and (2) the initialization method, which includes copying the original parameters or introducing noise.

\begin{table*}[ht]
  \centering
  \renewcommand{\arraystretch}{1.3} 
  \setlength{\tabcolsep}{8pt} 
  \footnotesize
  \scalebox{1.0}{
    \begin{tabular}{lccc}
      \toprule
      & No-noise & Gaussian ($\mu=0.01$) & Gaussian ($\mu=0.0001$) \\ 
      \midrule
    Among Layers & 73.1 & 43.1 & \cellcolor{lightblue} 72.8  \\ 
    After Model & 9.4  & 3.1 & 5.2 \\
      \bottomrule
    \end{tabular}
  }
  \caption{Layer extension initialization analysis. The original performance is 79.5.}
  \label{table:init}
\end{table*}

Table \ref{table:init} presents the initialization results for different layer extension methods. Our findings indicate that directly appending new layers to the model leads to a significant decline in performance, suggesting that abrupt structural modifications may disrupt the learned representations. In contrast, inserting new layers within the existing architecture introduces only a minor performance degradation, implying that gradual expansion is less disruptive to the model’s stability.  Additionally, we observe that duplicating layers without introducing noise achieves the highest performance, as it maintains the integrity of the original feature representations. On the other hand, adding Gaussian noise with a high mean substantially impacts performance, likely due to excessive perturbation of the initialized parameters. Initialization with all zeros preserves the original performance due to the residual connection. However, performance will degrade after training. Given that adding noise during initialization has the potential to improve training outcomes, we opt for an initialization method that applies Gaussian noise with a mean of $0.0001$, striking a balance between stability and adaptability.

\section{Model Training}

\subsection{Model Architecture}

Taking into account both accessibility and state-of-the-art performance, we select two model sizes: approximately 10B and 80B. Leveraging Qwen2.5B-7B and Qwen2.5-72B, we employ the model extension method described in Table \ref{table:extension} to initialize \texttt{Babel-9B} and \texttt{Babel-83B}. Specifically, as shown in Figure \ref{fig:layer}, we insert layers in the second half of the model, adding one every other layer.

\begin{table*}[ht]
  \centering
  \renewcommand{\arraystretch}{1.3} 
  \setlength{\tabcolsep}{8pt} 
  \footnotesize
  \scalebox{1}{
    \begin{tabular}{lcl}
      \toprule
      \textbf{Model} & \textbf{Initialization} & \textbf{Layer Inserting Position} \\ 
      \midrule
      Babel-9B  & \multirow{2}{*}{\begin{tabular}[c]{@{}l@{}}Duplicate + Gaussian Noise\end{tabular}} & \{14, 16, 18, 20, 22, 24\} \\ 
      Babel-83B &  & \{40, 42, 44, 46, 48, 50, 52, 54, 56, 58, 60, 62\} \\ 
      \bottomrule
    \end{tabular}
  }
  \caption{Layer extension method details.}
  \label{table:extension}
\end{table*}


\subsection{Pre-training Strategy}

\paragraph{Stage 1-Recovery.}

When we modify the parameters and disrupt the well-trained parameter collaboration, Babel's initial performance deteriorates compared to the original model (i.e., Qwen2.5 models). Consequently, in the first stage of pre-training, a large and diverse general training corpus encompassing all languages is crucial for recovery. Therefore, we sample a corpus for each language as equally as possible from the pre-training data, although achieving perfect equality can be challenging due to the limited availability of corpora for some languages. Additionally, to accelerate the performance recovery, we combine the English and Chinese training corpora during Stage 1 pre-training. For English, we leverage widely adopted, well-curated pre-training datasets such as RedPajama~\citep{weber2024redpajama} and Proof-Pile-2~\citep{paster2023openwebmath}, while for Chinese, we employ YAYI 2~\citep{luo2023yayi2multilingualopensource}.

\paragraph{Stage 2-Continuous Training.}

After recovery, the next step is to enhance multilingual capabilities, particularly for languages overlooked by previous models. To achieve this, we increase the proportion of low-resource languages in the pre-training corpus and continue training the model. In addition, we increase the proportion of textbooks in the training corpus, as tutorials are more effective for LLMs to acquire new knowledge.

\section{Evaluations}

We evaluate Babel against comparably sized open-source and commercial multilingual LLMs across a comprehensive set of multilingual tasks.

\subsection{Experiment Setup}

\paragraph{Dataset}
We employ multilingual tasks across several categories: (1) \textbf{World Knowledge}: MMMLU~\citep{openai2024mmmlu}, a human-translated version of MMLU~\citep{hendrycks2021measuringmassivemultitasklanguage} available in 14 languages. For languages not covered, we use Google Translate~\citep{GoogleTranslateAPI} to generate translations. Additionally, we include M3Exam~\citep{m3exam}, which consists of authentic human exam questions collected from various countries, covering multiple subjects and educational levels. (2) \textbf{Reasoning}: MGSM~\citep{shi2022languagemodelsmultilingualchainofthought} and XCOPA~\citep{ponti2020xcopamultilingualdatasetcausal}; (3) \textbf{Understanding}: XNLI~\citep{conneau2018xnlievaluatingcrosslingualsentence}; (4) \textbf{Translation}: Flore-200~\citep{nllbteam2022languageleftbehindscaling}.

\paragraph{Benchmark} 

For the 10B-size model, we compare Babel-9B with GLM4-9B~\citep{glm2024chatglm}, Gemma2-9B~\citep{gemmateam2024gemma2improvingopen}, Mistral-Nemo-2407\footnote{\url{https://huggingface.co/mistralai/Mistral-Nemo-Base-2407}} (referred to as Mistral-12B), Llama3.1-8B~\citep{dubey2024llama}, and Qwen2.5-7B~\citep{qwen2025qwen25technicalreport}, listed in order of their release dates. Furthermore, we compare Babel-83B with Llama3.1-70B~\citep{dubey2024llama} and Qwen2.5-72B~\citep{qwen2025qwen25technicalreport}.

\paragraph{Evaluation Details}

We utilize few-shot prompting methods across all datasets and models. For datasets other than Flore-200, accuracy serves as the evaluation metric, while for Flore-200, we use the chrF++ score, translating between each language and English.

\subsection{Main Results}


\begin{table*}[t]
\centering
\setlength{\tabcolsep}{4pt} 
\setlength{\extrarowheight}{2.4pt}
\scalebox{0.95}{ 
\begin{tabular}{lccccc|c} 
    \toprule
    \textbf{\normalsize{Dataset}}   & \textbf{\normalsize{GLM4-9B}} & \textbf{\normalsize{Gemma2-9B}} & \textbf{\normalsize{Mistral-12B}} &  \textbf{\normalsize{Llama3.1-8B}} & \textbf{\normalsize{Qwen2.5-7B}} & \textbf{\normalsize{Babel-9B}} \\
    \midrule
    MMMLU & 55.6 & \textbf{59.8} & 52.8 & 49.4 & 56.7 & 59.4  \\
    M3Exam & 56.6 & \textbf{61.6} & 54.2 & 52.5 & 58.8 & 61.3 \\
    XCOPA & 87.3 & 84.6 & 81.3 & 75.9 & 81.1 & \textbf{89.2} \\
    MGSM & 39.0 & 34.3 & 26.0 & 18.0 & 41.1 & \textbf{43.4}  \\
    XNLI & 69.9 & 61.7 & 55.0 & 48.9 & 70.3 & \textbf{71.9} \\
    Flores-200 & 46.6 & 53.2 & 50.8 & 50.9 & 45.5 & \textbf{55.1}  \\\midrule
    \rowcolor{lightblue}\textit{Average} & 59.2 & 59.5 & 53.4 & 49.3 & 58.9 & \textbf{63.4} \\
    \bottomrule
\end{tabular}
}
\caption{Performance of 10B-Size Base Models vs. Babel-9B-Base.}
\label{tab:multilingual_base}
\end{table*}

\begin{table*}[t]
\centering
\setlength{\tabcolsep}{4pt} 
\setlength{\extrarowheight}{2.4pt}
\scalebox{1.0}{ 
    \begin{tabular}{lcc|c} 
        \toprule 
  \textbf{\normalsize{Dataset}} & \textbf{\normalsize{Llama3.1-70B}} & \textbf{\normalsize{Qwen2.5-72B}} &  \textbf{\normalsize{Babel-83B}}   \\
        \midrule
        MMMLU & 69.1 & 74.7 & \textbf{76.3}    \\
        M3Exam & 67.4 & 71.2 & \textbf{72.1}  \\
        XCOPA & 92.6 & 81.1 & \textbf{92.8}  \\
        MGSM  & 48.9 & \textbf{63.9} & 62.6  \\
        XNLI & 66.2 & 74.9 & \textbf{76.6}  \\
        Flores-200 & 57.4 & 53.1 & \textbf{58.8}  \\\midrule
        \rowcolor{lightblue}\textit{Average} & 66.9 & 69.8 & \textbf{73.2} \\
        \bottomrule
    \end{tabular}
}
\caption{Performance of Open Large Multilingual LLMs vs. Babel-83B-Base.}
\label{tab:multilingual_80B_base}
\end{table*}

Table \ref{tab:multilingual_base} shows the results of Babel-9B-Base compared with 10B-size models. We find that Babel-9B-Base achieves the highest overall performance among the evaluated 10B-size base models, with an average score of 63.4, outperforming the closest competitor, Gemma2-9B-Base (59.5), by 3.9 points. Notably, Babel-9B-Base achieves the best results on XCOPA (89.2), MGSM (43.4), XNLI (70.9), and Flores-200 (55.1), demonstrating strong multilingual reasoning, understanding, and translation capabilities. While Gemma2-9B-Base performs competitively on MMMLU and M3Exam, Babel-9B-Base remains consistently strong across all benchmarks. Table \ref{tab:multilingual_80B_base} illustrates the results of Babel-83B-Base compared with open large multilingual LLMs. We find that Babel-83B achieves the highest overall performance among the evaluated models, with an average score of 73.2, outperforming the closest competitor, Qwen2.5-72B (69.8), by 3.4 points. Notably, Babel-83B achieves the best results on MMMLU (76.3), M3Exam (72.1), XCOPA (92.8), XNLI (76.6), and Flores-200 (58.8), demonstrating strong multilingual reasoning, understanding, and translation capabilities. These results highlight Babel’s effectiveness in multilingual understanding and reasoning, positioning it as the most capable open multilingual LLM within its parameter range.



\section{Further Analysis}

\subsection{Performance across Languages}

To further analyze Babel's performance across languages, we categorized them into high-resource and low-resource languages based on their scores in \citet{CommonCrawlLanguages}, a statistical measure derived from Common Crawl's monthly archives that reflects the availability of public training corpora. Languages with a score higher than 1 are classified as high-resource languages, including English, Chinese, German, Spanish, French, Indonesian, Italian, Japanese, Portuguese, Russian, and Vietnamese. In contrast, languages with a score lower than 1 are considered low-resource languages, including Hindi, Standard Arabic, Bengali, Urdu, Swahili, Tamil, Turkish, Korean, Javanese, Hausa, Thai, Iranian Persian, Filipino, and Burmese. We find that low-resource languages are those that have been underexplored by previous multilingual LLMs.

\begin{figure}[t]
    \centering
    \begin{subfigure}{0.32\textwidth}
        \centering
        \includegraphics[width=\textwidth]{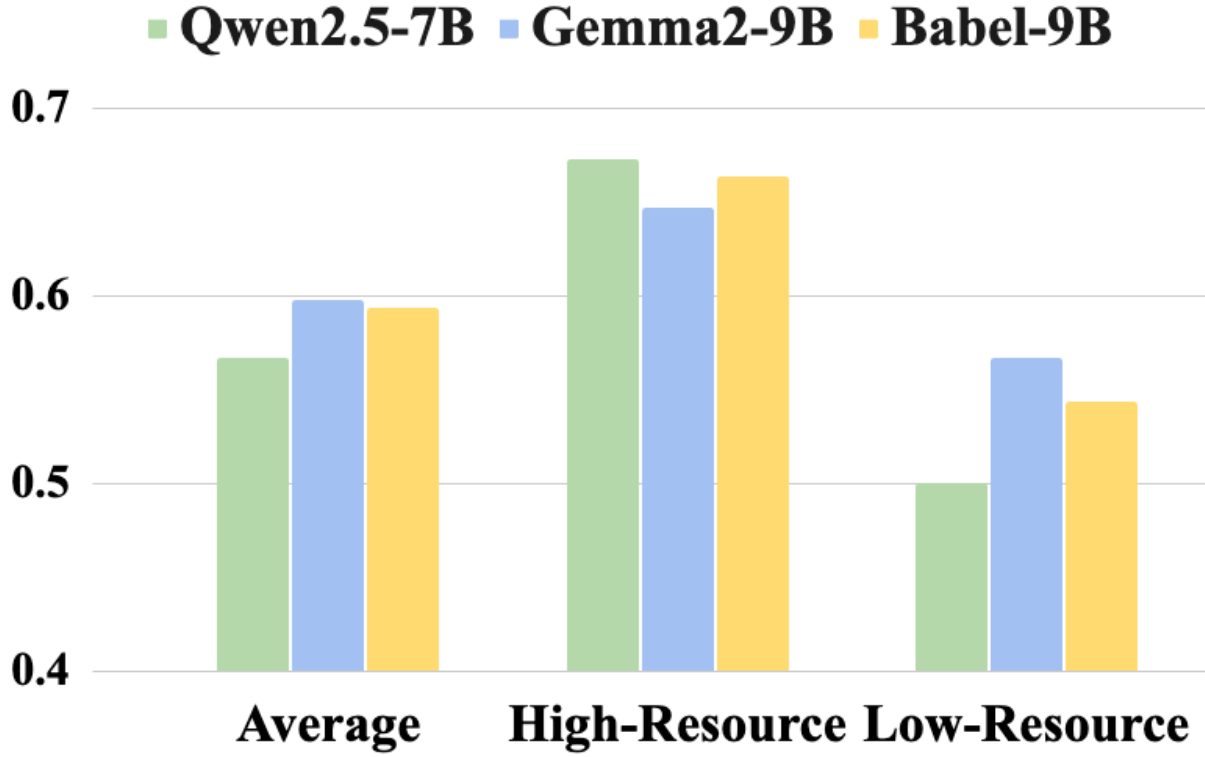}
        \caption{MMMLU}
        \label{fig:mmmlu}
    \end{subfigure}
    \hfill
    \begin{subfigure}{0.32\textwidth}
        \centering
        \includegraphics[width=\textwidth]{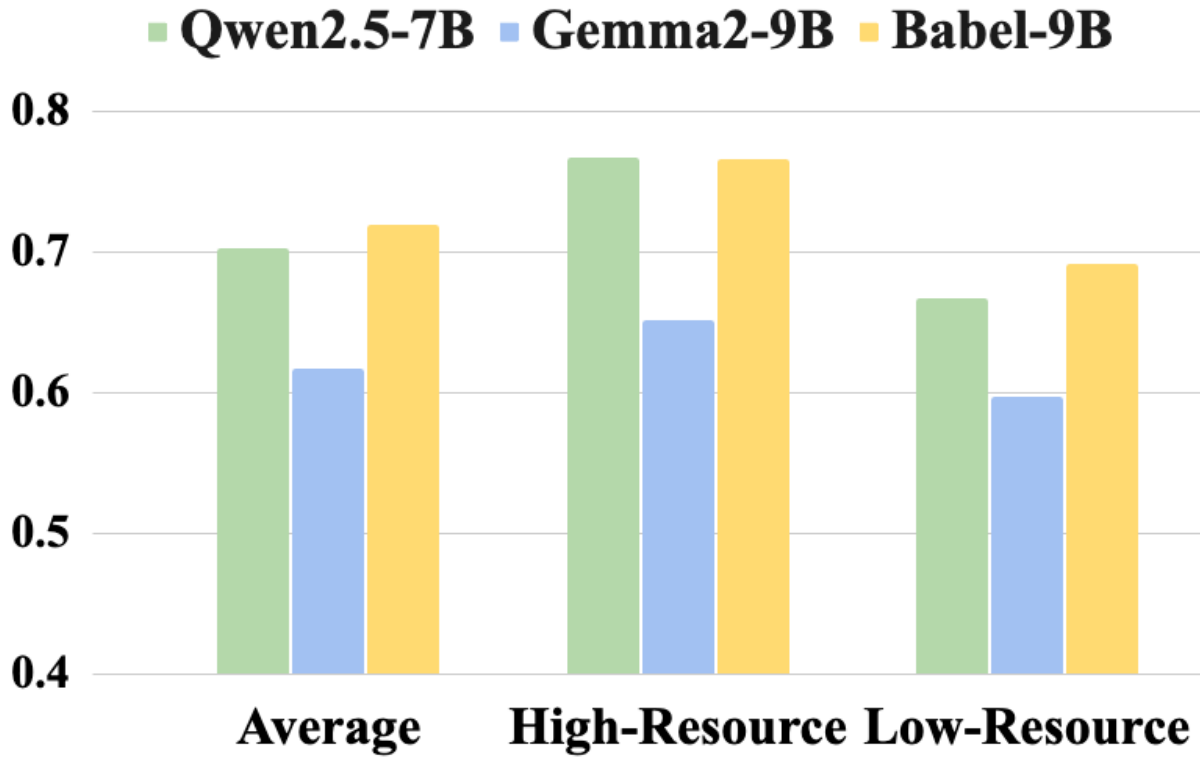}
        \caption{XNLI}
        \label{fig:m3exam}
    \end{subfigure}
    \hfill
    \begin{subfigure}{0.32\textwidth}
        \centering
        \includegraphics[width=\textwidth]{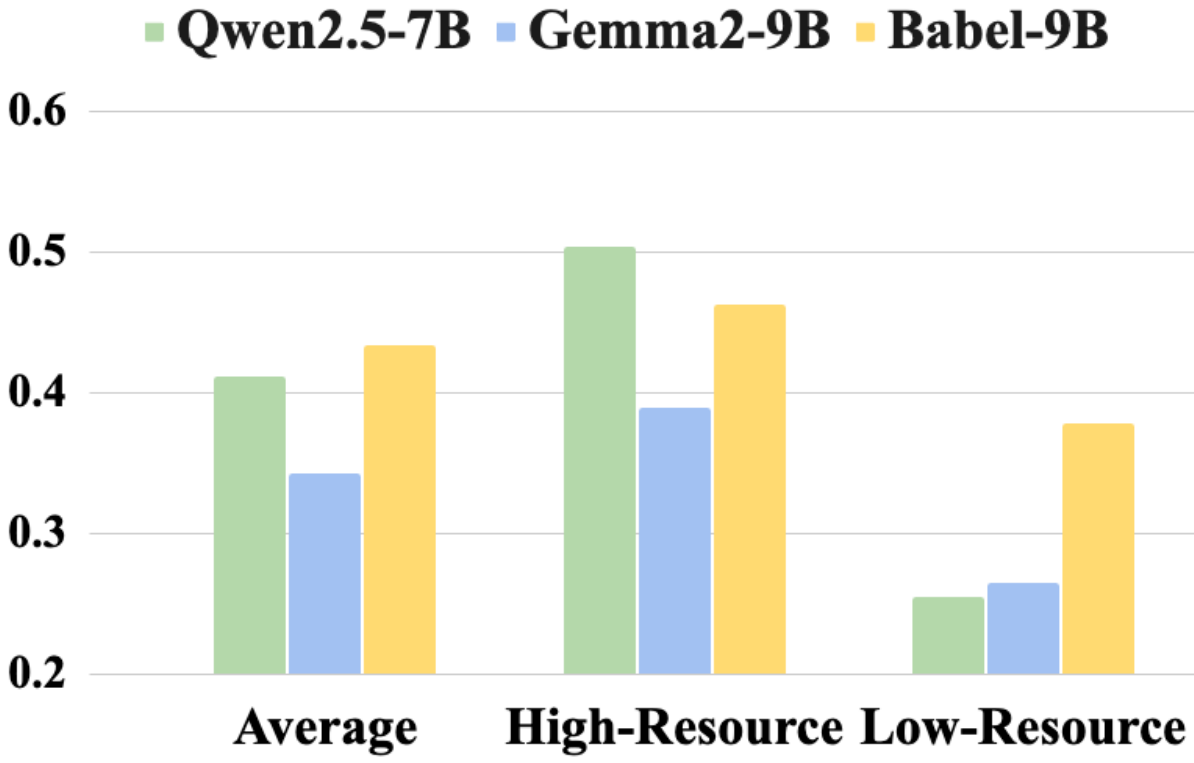}
        \caption{MGSM}
        \label{fig:mgsm}
    \end{subfigure}
    \caption{Performance of Babel-9B-Base comparison across languages.}
    \label{fig:combined}
\end{figure}

\begin{figure}[t]
\centering
\begin{minipage}{0.33\textwidth}
    \centering
    \setlength{\tabcolsep}{2pt} 
    \setlength{\extrarowheight}{4pt}
\scalebox{0.85}{ 
    \begin{tabular}{l|c|c}
    \toprule
    & \textbf{English} & \textbf{Multilingual} \\
    \midrule
    MMMLU   &  50.7  & \textbf{52.1} \\ 
    M3Exam  &  55.3   & \textbf{58.4}  \\ 
    XCOPA   &  84.2  & \textbf{83.3}  \\ 
    MGSM    &  41.8  & \textbf{42.1}  \\ 
    XNLI    &  64.5  & \textbf{67.8}  \\ 
    Flore-200 & 42.6 & \textbf{48.1}  \\ \midrule
    \rowcolor{lightblue} \textit{Average} & 56.5 & \textbf{58.6}  \\ 
    \bottomrule
    \end{tabular}}
    \captionof{table}{Performance comparison of English and multilingual SFT data.}
    \label{tab:example_table}
\end{minipage}%
\hfill
\begin{minipage}{0.65\textwidth}
    \centering
    \includegraphics[width=0.98\textwidth]{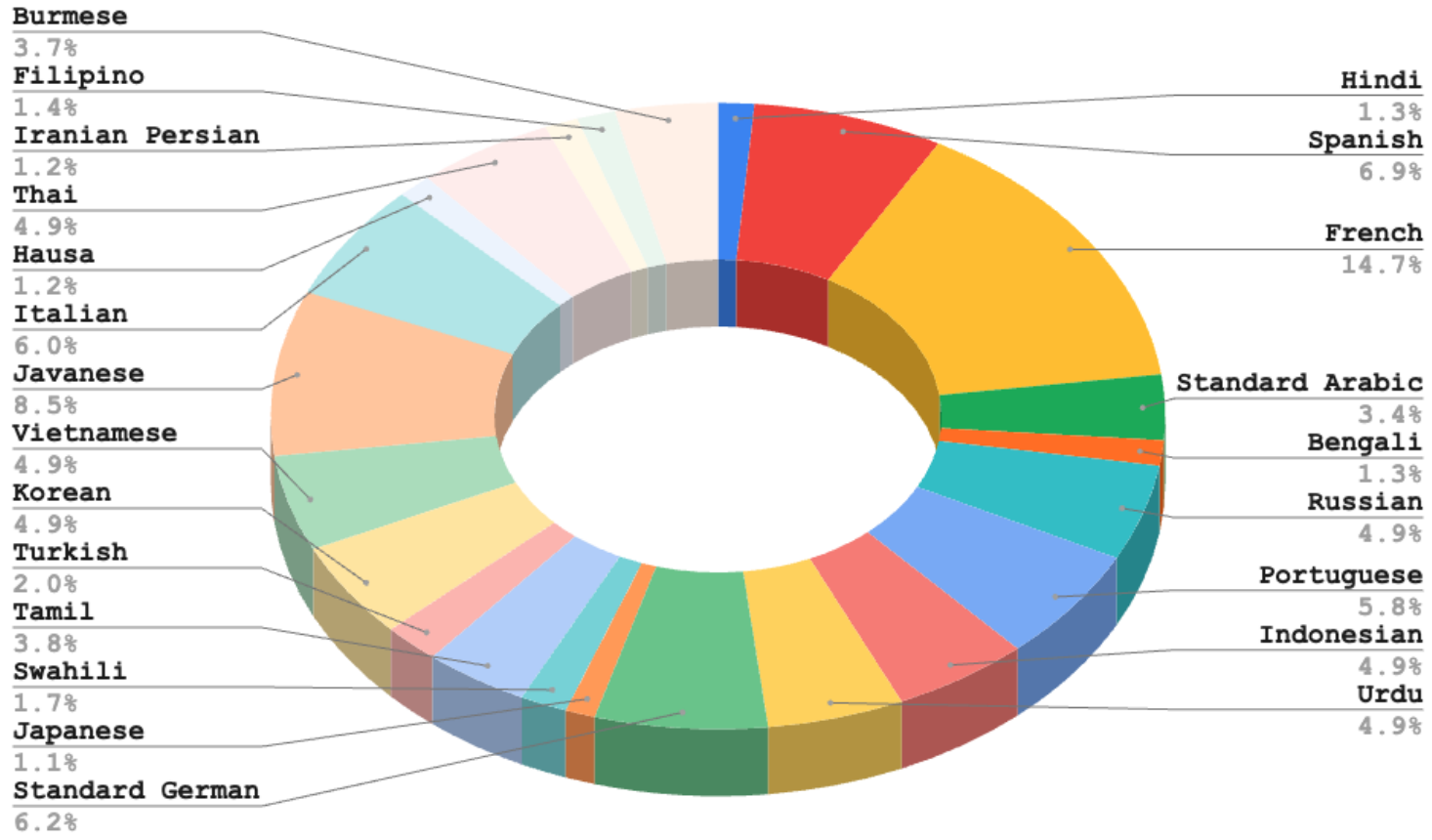}
    \captionof{figure}{Multilingual SFT data distribution excluding English and Chinese. }
    \label{fig:sft_stat}
\end{minipage}
\end{figure}

Figure \ref{fig:combined} illustrates the performance of Babel-9B-Base across high-resource and low-resource languages, compared to Qwen2.5-7B-Base and Gemma2-9B-Base. Qwen2.5-7B-Base serves as our backbone model, while Gemma2-9B-Base is the second most optimal multilingual LLM. Notably, Babel-9B-Base demonstrates significantly improved performance on low-resource languages compared to Qwen2.5-7B-Base (50.0 vs. 54.4 on MMMLU, 66.7 vs. 69.2 on XNLI, and 25.5 vs. 37.8 on MGSM). Conversely, when compared to Gemma2-9B-Base, Babel-9B-Base achieves higher performance on high-resource languages (64.7 vs. 66.4 on MMMLU, 65.2 vs. 76.6 on XNLI, and 38.9 vs. 46.3 on MGSM). Thus, Babel-9B-Base not only achieves the highest average performance but also strikes a balance between high-resource and low-resource languages.

\subsection{Supervised Fine-Tuning (SFT)}


\paragraph{SFT Data}

We primarily leverage open-source multilingual SFT training corpora and translated SFT training data. Specifically, we utilize WildChat~\citep{zhaowildchat}, a dataset comprising 1 million user-ChatGPT conversations with over $2.5$ million interaction turns. Additionally, we employ Everything Instruct Multilingual~\citep{everything_instruct_multilingual}, an extensive Alpaca-instruct-formatted dataset covering a diverse range of topics. 

Furthermore, we explore two approaches to constructing an SFT data pool with 400k conversations: one exclusively in English and the other in multiple languages. Table \ref{tab:example_table} compares these two SFT datasets, revealing that while English SFT data enhances the model’s instruction-following capability, multilingual SFT data yields significantly better overall performance. Consequently, we construct a larger multilingual SFT data pool.  Specifically, our final dataset consists of approximately 1 million multi-turn conversations. Figure \ref{fig:sft_stat} illustrates the distribution of SFT data across languages, excluding English and Chinese for better visualization. English comprises $40\%$ of the total SFT training data, while Chinese accounts for $10\%$.


\paragraph{SFT Training}

During training, conversations are packed together for efficiency, with a maximum token limit of 4096. The learning rate is configured to $4.0 \times 10^{-6}$, and a warmup ratio of 0.1 is applied.

\paragraph{Main Results}

\begin{table*}[t]
\centering
\setlength{\tabcolsep}{4pt} 
\setlength{\extrarowheight}{2.4pt}
\scalebox{0.9}{ 
\begin{tabular}{lccccc|c} 
    \toprule
    \textbf{\normalsize{Dataset}}   & \textbf{\normalsize{GLM4-9B}} & \textbf{\normalsize{Gemma2-9B}} & \textbf{\normalsize{Mistral-12B}} &  \textbf{\normalsize{Llama3.1-8B}} & \textbf{\normalsize{Qwen2.5-7B}} & \textbf{\normalsize{Babel-9B}} \\
    \midrule
    MMMLU & 53.9 & 59.6 & 52.0 & 50.6  & 56.0 & \textbf{59.8}  \\
    M3Exam & 55.0 & \textbf{63.2} & 54.1 & 54.2 & 58.0 & 62.9  \\
    XCOPA & 86.2 & 87.4 & 83.5 & 82.1 & 80.4 & \textbf{88.9} \\
    MGSM  & 52.2 & 62.4 & 41.4 & 37.2 & 59.1 & \textbf{64.3} \\
    XNLI & 66.2 & 66.7 & 56.1 & 55.8 & 68.3 & \textbf{72.4} \\
    Flores-200 & 50.8 & 54.8 & 48.9 & 47.3 & 45.8 & \textbf{56.7}  \\\midrule
    \rowcolor{lightblue}\textit{Average} & 60.7 & 65.7 & 56.0 & 54.5 & 61.3 & \textbf{67.5} \\
    \bottomrule
\end{tabular}
}
\caption{Performance of 10B-Size Instruct Models vs. Babel-9B-Chat}
\label{tab:multilingual_chat}
\end{table*}

\begin{table*}[t]
\centering
\setlength{\tabcolsep}{4pt} 
\setlength{\extrarowheight}{2.4pt}
\scalebox{1.0}{ 
    \begin{tabular}{l>{\columncolor{codegray}}ccc|c} 
        \toprule 
  \textbf{\normalsize{Dataset}} & \textbf{\normalsize{GPT-4o}}  & \textbf{\normalsize{Qwen2.5-72B}} & \textbf{\normalsize{Llama3.1-70B}} &  \textbf{\normalsize{Babel-83B}}  \\
        \midrule
        MMMLU  & 77.3 & 73.0 & 71.7 & \textbf{76.8}   \\
        M3Exam & 74.9 & 70.2 & 69.5 & \textbf{73.2}  \\
        XCOPA & 90.6 & 89.2 & 92.2 & \textbf{92.7}  \\
        MGSM & 83.1 & \textbf{75.8} & 56.7 & 72.5 \\
        XNLI & 69.6 & 72.6 & 55.8 & \textbf{76.3} \\
        Flores-200  & 54.9 & 50.4 & \textbf{56.1} & 54.8  \\\midrule
        \textit{Average} & 75.1 & 71.9 & 67.0 & \textbf{74.4} \\
        \bottomrule
    \end{tabular}
}
\caption{Babel-83B-Chat vs. Leading Open Multilingual LLMs and the Top Commercial Model. Results for the best open multilingual models are \textbf{bolded}.}
\label{tab:multilingual_80B}
\end{table*}

Table~\ref{tab:multilingual_chat} shows that Babel-9B-Chat achieves the highest average score (67.5), surpassing Gemma2-9B-Instruct (65.7) and other models. It leads in five out of six benchmarks, excelling in XCOPA (88.9), MGSM (64.3), XNLI (72.4), and Flores-200 (56.7), demonstrating strong multilingual reasoning and problem-solving. While Gemma2-9B-Instruct slightly outperforms it on M3Exam (63.2 vs. 62.9), Babel-9B-Chat remains consistently strong across tasks. Table \ref{tab:multilingual_80B} illustrates that Babel-83B-Chat achieves the highest average performance (74.4) among open multilingual LLMs, closely trailing GPT-4o (75.1) and outperforming Qwen2.5-72B-Instruct (71.9) and Llama3.1-70B-Instruct (67.0). With leading scores in XCOPA and XNLI, Babel-83B-Chat is the closest open multilingual LLM to the best commercial alternative, demonstrating strong multilingual capabilities.

Note that results are achieved purely by leveraging publicly available datasets, showcasing the robust foundational performance of Babel base models. We believe that incorporating more SFT data across diverse types, domains, and formats, along with additional alignment data and preference tuning, will further enhance the chat version beyond its current capabilities.




\section{Conclusion}


We introduced Babel, an open multilingual LLM that addresses the gap in language coverage by supporting 25 widely spoken languages, including under-resourced ones, and covering over 90\% of the global population. Using a novel layer extension technique, Babel achieves state-of-the-art performance, with its two variants, Babel-9B and Babel-83B, excelling in multilingual benchmarks. By setting new standards for open multilingual LLMs, Babel underscores the importance of inclusivity in NLP development and sets a strong foundation for future research in multilingual language modeling.

\section*{Acknowledgments}

We would like to thank Guanzheng Chen for assisting with the implementation of the training codebase. Our special thanks go to our professional and native linguists—Tantong Champaiboon, Nguyen Ngoc Yen Nhi, and Tara Devina Putri—who contributed to building, evaluating, and fact-checking our sampled pretraining dataset. We also appreciate Fan Wang, Jiasheng Tang, Xin Li, and Hao Zhang for their efforts in coordinating computing resources.

\bibliography{anthology,custom}
\bibliographystyle{acl_natbib}





\end{document}